%% file: main.tex
\documentclass[11pt]{article}

\usepackage[final]{acl}

\usepackage{times}
\usepackage{latexsym}
\usepackage{multirow}
\usepackage{subcaption}
\usepackage{booktabs}
\usepackage{pifont}
\usepackage{amssymb}
\usepackage{threeparttable}
\usepackage{booktabs}
\usepackage{amsmath}

\usepackage[T1]{fontenc}

\usepackage[utf8]{inputenc}

\usepackage{microtype}

\usepackage{inconsolata}

\usepackage{graphicx}

\title{River-LLM: Large Language Model Seamless Exit Based on KV Share}

\author{Yingtao Shen \\
  Shanghai Jiao Tong University \\
  Shanghai, China  \\
  \texttt{doctorcoal@sjtu.edu.cn} \\\And
  An Zou \\
  Shanghai Jiao Tong University \\
  Shanghai, China \\
  \texttt{an.zou@sjtu.edu.cn} \\}

\begin{document}
\maketitle
\begin{abstract}
Large Language Models (LLMs) have demonstrated exceptional performance across diverse domains but are increasingly constrained by high inference latency. Early Exit has emerged as a promising solution to accelerate inference by dynamically bypassing redundant layers. However, in decoder-only architectures, the efficiency of Early Exit is severely bottlenecked by the KV Cache Absence problem, where skipped layers fail to provide the necessary historical states for subsequent tokens. Existing solutions, such as recomputation or masking, either introduce significant latency overhead or incur severe precision loss, failing to bridge the gap between theoretical layer reduction and practical wall-clock speedup. In this paper, we propose River-LLM, a training-free framework that enables seamless token-level Early Exit. River-LLM introduces a lightweight KV-Shared Exit River that allows the backbone's missing KV cache to be naturally generated and preserved during the exit process, eliminating the need for costly recovery operations. Furthermore, we utilize state transition similarity within decoder blocks to predict cumulative KV errors and guide precise exit decisions. Extensive experiments on mathematical reasoning and code generation tasks demonstrate that River-LLM achieves $1.53\times$ to $2.16\times$ practical speedup while maintaining high generation quality.

\end{abstract}

\input{latex/intro}

\input{latex/related}

\input{latex/motivation}

\input{latex/method}

\input{latex/evaluation}

\input{latex/conclusion}

\newpage
\section*{Limitations}
\begin{itemize}
    \item Our current evaluation focuses on representative models up to 8B parameters. While the framework is designed to be architecture-agnostic, further validation on larger scales, such as 24B and 70B models, is necessary to confirm its performance and scalability at extreme parameter counts.
    \item As River-LLM is primarily optimized for token-level autoregressive decoding, its efficiency gains are most significant during the generation phase. Consequently, the speedup is less pronounced for prefill-dominant tasks, such as those within the MMLU benchmark, where a sequence-level exit strategy is currently applied.
\end{itemize}

\section*{Acknowledgments}
This work was supported in part by the the Open Research Fund of Peng Cheng Laboratory under Grant 2025KF1B0010. An Zou is the corresponding author.

The authors acknowledge the use of Gemini 3.0 Pro for polishing assistance. This assistance was limited to polishing, and all research findings and final content were independently verified by the authors.

\bibliography{main}

\appendix

\input{latex/appendix}

\end{document}

%% file: latex/intro.tex
\section{Introduction}
\label{sec:intro}

In recent years, Large Language Models (LLMs), particularly decoder-only language models, have driven transformative advancements globally, demonstrating expert-level performance in domains such as code generation \cite{roziere2023code}, complex reasoning \cite{wei2022chain}, and creative writing \cite{bubeck2023sparks}. Despite their remarkable capabilities, the enormous parameter counts of LLMs impose significant challenges, including substantial inference latency and excessive hardware energy consumption \cite{pope2023efficiently}.

To mitigate these costs, dynamic inference \cite{han2021dynamic} has emerged as a promising paradigm for optimizing token generation. Among various techniques, Early Exit, originally proven effective in classical CNNs \cite{branchynet, conditional}, is gaining traction as a dominant trend for LLM acceleration. By dynamically bypassing redundant layers based on input complexity, Early Exit enables sample-dependent computation reduction. Recent advancements encompass sophisticated skipping strategies \cite{SkipDecode, DiffSkip}, specialized exit layer designs \cite{EE-LLM, Balcony}, and optimizations for specific reasoning modes such as Chain-of-Thought \cite{DEER} or speculative drafting \cite{LayerSkip}.

\begin{figure}[!t]
\centering
\includegraphics[width=0.48\textwidth]{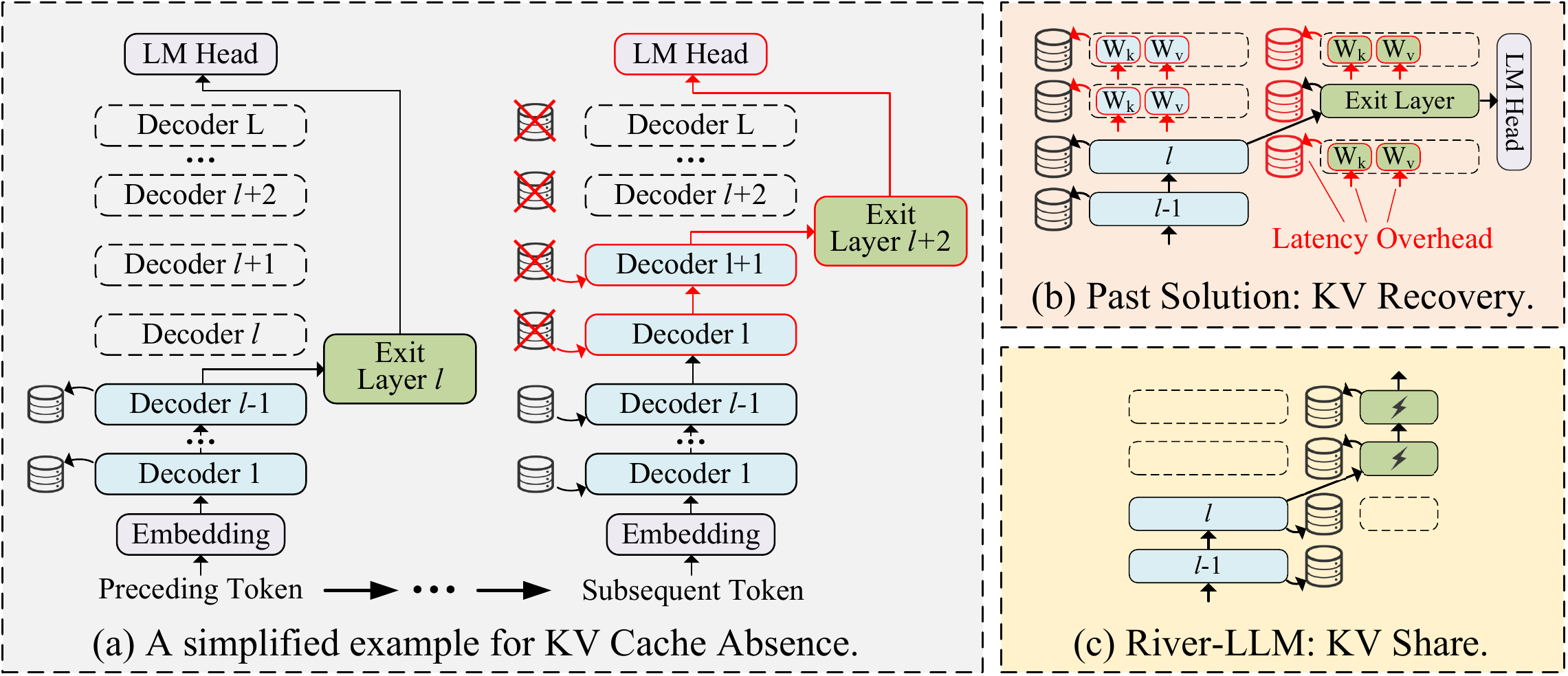}
\caption{KV Cache Absence problem for Early Exit in decoder-only LLM.}
\label{fig:absence}
\end{figure}

However, a critical gap exists between the theoretical potential and practical efficiency of Early Exit in LLMs. Our empirical analysis reveals that while over 50\% of tokens can theoretically exit at early layers without compromising output consistency, the actual wall-clock speedup during autoregressive inference remains marginal. This discrepancy is primarily attributed to the \textbf{KV Cache Absence} problem. As illustrated in Fig. \ref{fig:absence}, when a token exits early, it fails to compute the Key-Value (KV) pairs for the bypassed layers. Consequently, subsequent tokens lack the necessary history for their own self-attention computations at those layers. Existing remedies, such as Batching Recompute \cite{EE-LLM}, State Propagation \cite{CALM}, or KV Masking \cite{D-LLM}, either introduce significant latency overhead or suffer from severe accuracy degradation. None of these approaches fully resolve the conflict between token-level exiting and KV cache integrity.

Inspired by recent explorations in KV cache redundancy \cite{liu2024minicache}, we propose \textbf{River-LLM}, a framework that achieves \textbf{seamless token-level exit} by constructing a "KV-Shared Exit River". River-LLM eliminates the need for costly recovery operations by ensuring that the skipped backbone layers' KV information is naturally preserved or approximated through a lightweight, shared structure. Our primary contributions are summarized as follows:

\begin{itemize}
    \item We provide a systematic evaluation of existing methods for addressing KV Cache Absence. By establishing a unified testing environment, we demonstrate that current strategies varying from masking to recomputation fail to bridge the gap between theoretical layer reduction and practical latency savings.
    \item We introduce the first "seamless exit" mechanism for LLMs. By designing a lightweight KV-Shared Exit Layer, River-LLM allows the backbone’s missing KV cache to be implicitly generated during the exit process. Furthermore, we utilize the similarity between decoder inputs and outputs to predict cumulative KV errors, guiding precise exit decisions.
    \item Experimental results across diverse long-sequence tasks, including mathematical reasoning and code generation, demonstrate that River-LLM achieves $1.53\times$ to $2.16\times$ wall-clock speedup without requiring any additional training or fine-tuning.
\end{itemize}

%% file: latex/related.tex
\section{Related Works}
\label{sec:related}


Early Exit is a cornerstone of dynamic neural networks designed to reduce computational redundancy \cite{han2021dynamic}. Initially, the Early Exit methods for Large Language Models (LLMs) was inherited from the previous Early Exit work on other neural network architectures. For instance, DAT \cite{DAT} introduced a variable-layer Transformer reminiscent of the BranchyNet architecture \cite{branchynet, conditional}. Subsequent works like CALM \cite{CALM} explored confidence metrics tailored for LLMs, while AdaInfer \cite{AdaInfer} is similar to \cite{li2023predictive}, it trains an SVM to predict at which layer the LLM can achieve results consistent with those of the last layer.

As decoder-only architectures became dominant, research shifted toward leveraging the unique structural and inferencing properties of these models. One primary direction focuses on skipping strategies. SkipDecode \cite{SkipDecode} bypasses intermediate decoders based on a pre-defined computational budget, while DiffSkip \cite{DiffSkip} and AdaSkip \cite{AdaSkip} utilize similarity between attention vectors to make on-the-fly skipping decisions. Notably, D-LLM \cite{D-LLM} incorporates MLPs to control layer skipping and introduces a mask mechanism to maintain KV Cache consistency. RAEE \cite{RAEE} further enhances control by retrieving exit distributions from similar tasks. Parallelly, another research thread focuses on exit layer design to mitigate accuracy loss. Balcony \cite{Balcony} transfers the final layer of the backbone as fine-tuned exit layers, whereas EE-LLM \cite{EE-LLM} provides a diverse library of exit modules (MLP, Decoder, etc.) and implements parallel GPU scheduling to minimize overhead.

Beyond general autoregressive generation, Early Exit has recently been integrated into specialized applications. LayerSkip \cite{LayerSkip} and SpecEE \cite{SpecEE} combine Early Exit with Speculative Decoding to accelerate the drafting stage, pushing the Pareto frontier of inference efficiency. Furthermore, DEER \cite{DEER} adapts the Early Exit concept to Chain-of-Thought (CoT) reasoning by truncating the reasoning chain based on confidence.

In this work, we distinguish Early Exit from other dynamic inference paradigms such as Router-Tuning \cite{Router-Tuning} and MoNE \cite{MoNE}. While those methods treat dynamic routing as an intrinsic architectural property, Early Exit focuses on losslessly or near-losslessly bypassing computations within a fixed backbone. Consequently, we categorize "skip-based" and "dynamic depth" approaches under the broader umbrella of Early Exit.


%% file: latex/motivation.tex
\section{Motivation}
\label{sec:motivation}
In this section, we first investigate the full potential of the Early Exit technique in Large Language Models. A series of experiments conducted on the Llama3.2 1B model with pre-trained decoder-based exit layer will demonstrate that the finest-granularity Early Exit can achieve a theoretical inference speedup of up to $3.3\times$. Following this, we will delve into why existing Early Exit techniques have been largely confined to the academic domain so far. Experiments will elucidate the critical challenge posed by KV Cache Absence, which creates a dilemma between autoregressive inference quality and KV cache recovery overhead. Existing methods often compromise between these two factors, leading to a situation where the promised theoretical speedup often fails to materialize under practical inference frameworks.

\begin{figure}[t]
\centering
	\begin{subfigure}[b]{0.235\textwidth}
	\includegraphics[height=0.098\textheight]{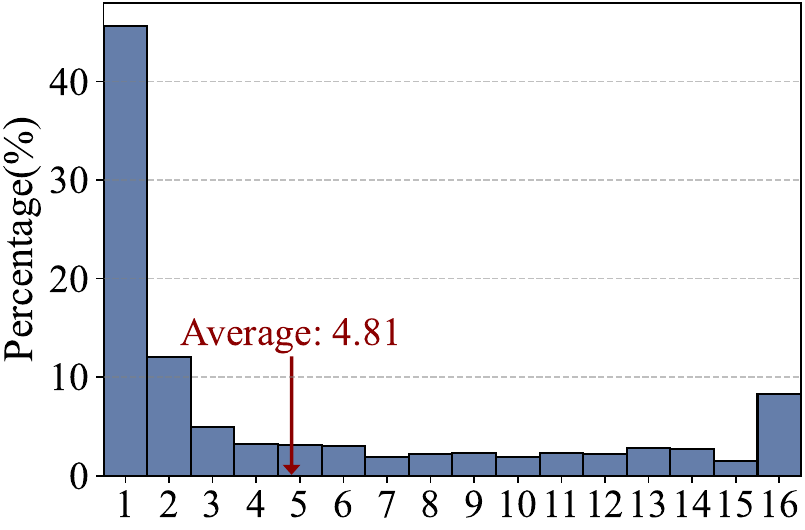}
    \caption{}
	\label{fig:dist}
	\end{subfigure}
	\hfill
	\begin{subfigure}[b]{0.235\textwidth}
	\centering
	\includegraphics[height=0.10\textheight]{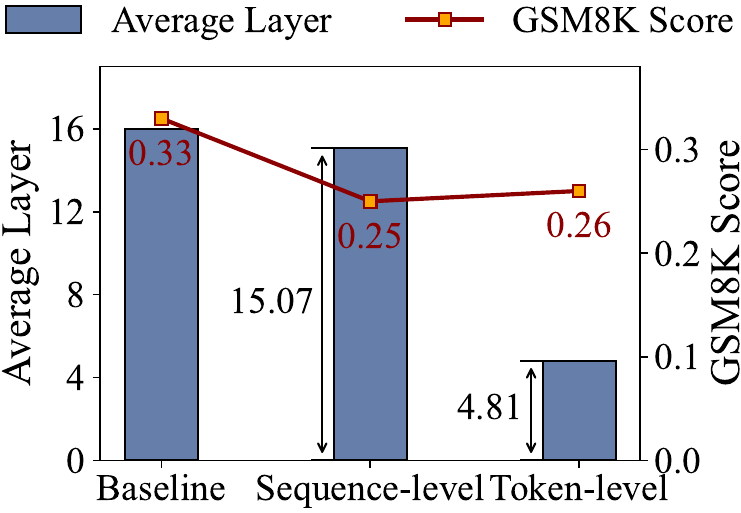}
    \caption{}
	\label{fig:layer}
	\end{subfigure}
\label{fig:token-level}
\caption{(a) Distribution of optimal Token-level Exit position for Llama3.2 1B on GSM8K. $Score=0.26$. (b) Token-level Exit significantly outperforms Sequence-level Exit on GSM8K.}
\end{figure}


\subsection{Sequence-level Exit and Token-level Exit}
\label{sec:token-level}

Two distinct exit granularities exist when applying Early Exit to Large Language Models: Sequence-level Exit and Token-level Exit. For a single autoregressive generation step in a decoder-only model, Sequence-level Exit mandates a fixed exit position for every token within that sequence. For instance, LayerSkip \cite{LayerSkip} directly employs a fixed layer $E$ to specify the stopping layer for the Draft Model during the entire generation process. Balcony \cite{Balcony} is even more straightforward, using Balcony-L/M/S to denote inference modes that exit at $\frac{3}{4}$, $\frac{1}{2}$, and $\frac{1}{4}$ of the model depth, respectively. The primary advantage of Sequence-level Exit is its simplicity of implementation and direct portability from previous Early Exit works (e.g., BranchyNet \cite{branchynet}). However, the exit layer for the entire sequence must be determined by the latest exit token, which leads to significant lost opportunities for early termination during long sequence generation. While Sequence-level Exit still offers acceleration in specific contexts, such as when LayerSkip \cite{LayerSkip} and SpecEE \cite{SpecEE} integrate Early Exit with Speculative Decoding for early termination on a limited-length draft token sequence. These methods only demonstrate efficacy in constrained scenarios, and their scalability remains highly limited.

In contrast, Token-level Exit has emerged as the prevailing paradigm for decoder-only LLMs due to its superior granularity. For instance, SkipDecode \cite{SkipDecode} explicitly highlights its "token-level" nature as a core feature. This approach allows each token in an autoregressive sequence to exit at its own optimal depth. To quantify the potential computational gains, we profile the optimal exit position for each token (defined as the the shallowest exit layer yielding the same prediction as the final layer) using Llama3.2 1B on the GSM8K benchmark with batch size of 1. Fig.~\ref{fig:dist} illustrates the distribution of these optimal exit layers, while Fig.~\ref{fig:layer} compares the average layers executed by Token-level and Sequence-level Exit at comparable accuracy levels. Despite the inherent difficulty of GSM8K for a model of this scale, Token-level Exit achieves a theoretical speedup of $3.3\times$ with 6\% accuracy drop, assuming zero overhead for KV Cache recovery and exit decision logic. Evaluations on the GSM8K benchmark (averaging over 100 tokens per response) underscore that \textbf{Token-level Exit possesses potential far exceeding that of Sequence-level Exit, especially for long-sequence generation}. However, the unique autoregression inference of decoder-only models introduces a critical bottleneck for Token-level Exit: KV Cache Absence.


\subsection{Key Challenge: KV Cache Absence for Decoder-only Model}
\label{sec:KV absence}
As illustrated in Figure~\ref{fig:absence}, the KV Cache is one of the core mechanisms for autoregressive inference in decoder-only language models. While KV Cache prevents the redundant recomputation of keys and values for preceding tokens during each generation step, it consequently creates a data dependency between the generation process of the current token and that of previous tokens.
An early exit of a preceding token directly leads to the absence of KV values from the skipped decoder layers (termed KV Cache Absence). This subsequently prevents the forward pass of later tokens that rely on these layers from accessing the necessary prior Keys and Values. KV Cache Absence is the fundamental challenge distinguishing LLM Early Exit from other traditional neural networks.
Almost all Token-level Exit works acknowledge this problem and attempt to address it. These existing strategies can be broadly categorized into four types:

\begin{figure}[t]
\centering
	\begin{subfigure}[b]{0.235\textwidth}
	\includegraphics[height=0.125\textheight]{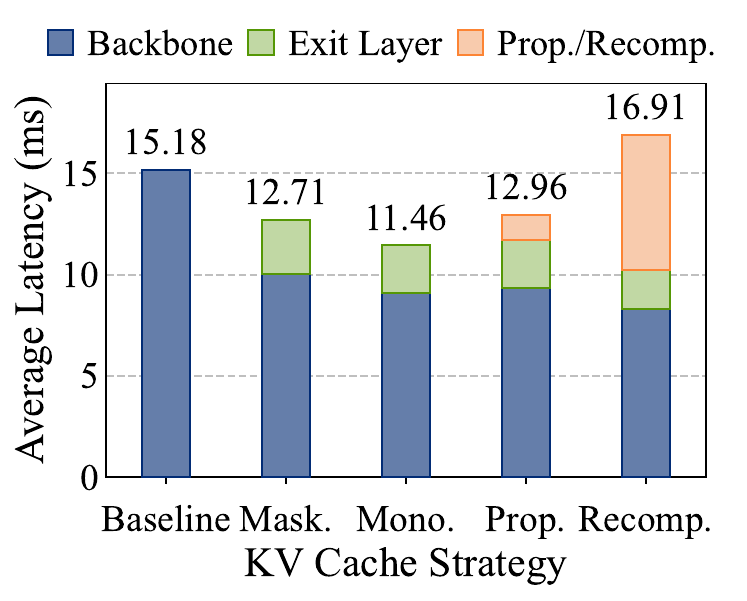}
    \caption{}
	\label{fig:latency1}
	\end{subfigure}
	\hfill
	\begin{subfigure}[b]{0.235\textwidth}
	\centering
	\includegraphics[height=0.125\textheight]{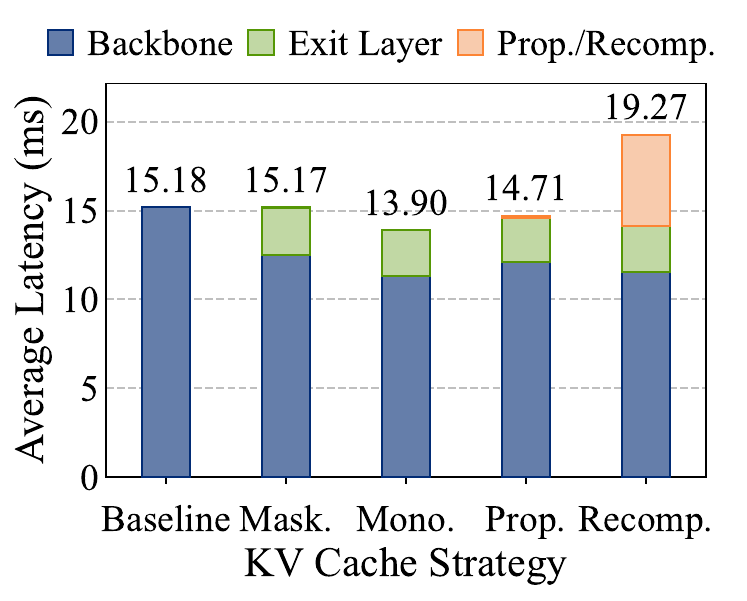}
    \caption{}
	\label{fig:latency2}
	\end{subfigure}
\caption{Average $ms/token$ of Token-level Exit using difference KV Cache Strategy on GSM8K. (a) Relaxed threshold, $Score\approx0.15$. (b) Strict threshold, $Score\approx0.25$.}
\label{fig:latency}
\end{figure}

\begin{itemize}
    \item \textbf{Batching Recompute}: The principle of KV Recompute is straightforward: when the layer about to be executed lacks the KV Cache of a preceding token, the missing KV Cache is recalculated. Early Exit methods based on Batching Recompute typically cache the hidden state before exit. If the forward pass encounters KV Cache Absence, the previously saved hidden state is used to resume the inferencing until the necessary KV data is satisfied. While this approach reduces memory footprint of key and values, it offers negligible computational savings during long sequence generation and requires extra hardware batching support. For instance, EE-LLM \cite{EE-LLM} utilizes model parallelism and scheduling mechanisms to allow KV Recompute to run as a parallel cudastream, leveraging idle GPU resources to mitigate its overhead. 
    \item \textbf{Mono-Decreasing Exit}: Mono-Decreasing Exit circumvents the KV Cache problem by regulating the token exit positions. For example, SkipDecode \cite{SkipDecode} strictly enforces that the exit position of every token within a single sequence adheres to a monotonically decreasing constraint to meet the KV Cache requirements of subsequently generated tokens. At the cost of restricting exit opportunities, Mono-Decreasing Exit offers a simple solution to avoid the KV Cache Absence issue.
    \item \textbf{State Propagation}: State Propagation is a classic Early Exit KV strategy, directly adopted by DAT \cite{DAT}, CALM \cite{CALM}, and ELLM \cite{ELLM}. In contrast to KV Recompute, State Propagation copies the hidden states from the current exit layer to serve as input for every subsequent decoder layer, directly computing the KV Cache for those skipped layers. This design represents a trade-off that balances the impact of KV Cache recovery on generation quality and inference speed.
    \item \textbf{KV Mask}: KV Mask is an aggressive strategy that modifies the original Causal Mask mechanism to make all tokens with missing KV Cache invisible to the current decoder layer. This naturally leads to a noticeable impact on generation quality. To compensate, D-LLM \cite{D-LLM} places a constraint on the minimum allowed exit layer to ensure limited quality degradation.
\end{itemize}

We evaluated the average per-token latency of these strategies on Llama3.2 1B at two consistent accuracy level, with results summarized in Fig.~\ref{fig:latency}. The KV Mask approach exhibits the highest backbone latency, as it necessitates executing deeper layers to compensate for the precision degradation caused by missing KV values. While KV Recompute involves fewer backbone layers, it incurs substantial computational costs and disrupts memory access efficiency during long-sequence generation without specialized multi-GPU scheduling. State Propagation offers an approximation that trades off accuracy for overhead; however, its practical speedup remains inferior to the Mono-Decreasing Exit, which achieves better latency at the cost of restricting exit flexibility.

\newcommand{\cmark}{\textcolor[rgb]{0.483,0.706,0.424}{$\checkmark$}}
\newcommand{\xmark}{\textcolor[rgb]{0.722,0.329,0.314}{\ding{55}}}

\begin{table*}[t]
\begin{footnotesize}
\setlength{\tabcolsep}{3pt}
  \centering
  \caption{Decoder-only transformer early exit techniques.}
    \begin{tabular}{cccccc}
    \toprule
    \textbf{Exit Granularity} & \textbf{KV Cache Strategy} & \textbf{Representative Methods} & \textbf{Seamless Exit} & \textbf{Train} & \textbf{Latency} \\
    \midrule
    \multirow{2}[1]{*}{Sequence Level} & None (Independent) & Balcony & \textcolor[rgb]{ 1,  0,  0}{No} & \textcolor[rgb]{ 1,  0,  0}{Train Exit Layer} & \textcolor[rgb]{ 1,  0,  0}{High} \\
            & Speculation KV Reuse & Layerskip, SpecEE & \textcolor[rgb]{ 1,  0,  0}{No} & \textcolor[rgb]{ 1,  0,  0}{Train Draft Model} & \textcolor[rgb]{ 1,  .753,  0}{Medium} \\
    \midrule
    Limited Token Level & Mono-Decreasing Exit & SkipDecode & \textcolor[rgb]{ 1,  0,  0}{No} & \textcolor[rgb]{ 1,  0,  0}{Train Skip Predictor} & \textcolor[rgb]{ 1,  .753,  0}{Medium} \\
    \midrule
    \multirow{4}[1]{*}{Token Level} & Batching Recompute & EE-LLM  & \textcolor[rgb]{ 1,  0,  0}{No} & \textcolor[rgb]{ 1,  0,  0}{Train Exit Layer} & \textcolor[rgb]{ 1,  0,  0}{High} \\
            & State Propagation & ELLM, CALM & \textcolor[rgb]{ 1,  0,  0}{No} & \textcolor[rgb]{ 1,  0,  0}{Train Exit Layer} & \textcolor[rgb]{ 1,  .753,  0}{Medium} \\
            & KV Mask & D-LLM   & \textcolor[rgb]{ 1,  0,  0}{No} & \textcolor[rgb]{ 1,  0,  0}{Train Skip Predictor} & \textcolor[rgb]{ 1,  .753,  0}{Medium} \\
            & \textbf{KV Share} & \textbf{River-LLM (Ours)} & \textcolor[rgb]{ 0,  .69,  .314}{\textbf{Yes}} & \textcolor[rgb]{ 0,  .69,  .314}{\textbf{Train-Free}} & \textcolor[rgb]{ 0,  .69,  .314}{\textbf{Low}} \\
    \bottomrule
    \end{tabular}%
  \label{tab:seamless}%
\end{footnotesize}
\end{table*}%

\textbf{In summary}, none of these strategies fully eliminates the penalty of KV Cache Absence. Neglecting KV integrity leads to severe performance degradation; recomputation significantly hampers the net acceleration; and imposing exit constraints severely limits the inherent potential of Token-level Exit. These overheads and limitations collectively obstruct the practical realization of early exit benefits.

To address these challenges, we propose River-LLM, a \textbf{seamless} exit framework. We define an LLM Early Exit mechanism as "\textbf{seamless}" if it inherently achieves:
\begin{itemize}
    \item \textbf{Granular Freedom}: Individual tokens can exit at arbitrary layers independently.
    \item \textbf{Intrinsic KV Integrity}: The KV cache for skipped layers is automatically populated as a byproduct of the exit path's execution, eliminating the need for post-exit recovery or recomputation.
\end{itemize}
Table~\ref{tab:seamless} shows the core advantages of River-LLM over previous methods. Built upon the principle of KV Sharing, River-LLM enables tokens to traverse an "Exit River" rapidly regardless of their exit point. Crucially, the exit layers naturally substitute for the skipped decoder layers to generate a complete KV cache without incurring additional operational overhead.

%% file: latex/method.tex
\section{Seamless Exit: River-LLM}
\label{sec:river}
River-LLM is designed as a scalable and architecture-agnostic framework for early exit, applicable to various decoder-only Transformer models. This section first introduces the KV-Shared Exit Layer, the core component responsible for maintaining Intrinsic KV Integrity. We then detail how these modules are orchestrated during inference to achieve maximum Granular Freedom and substantial acceleration.

\subsection{KV-Shared Exit Layer}
\label{sec:kv_share}
\begin{figure*}[!t]
\centering
	\begin{subfigure}[c]{0.53\textwidth}
    \centering
	\includegraphics[height=0.16\textheight]{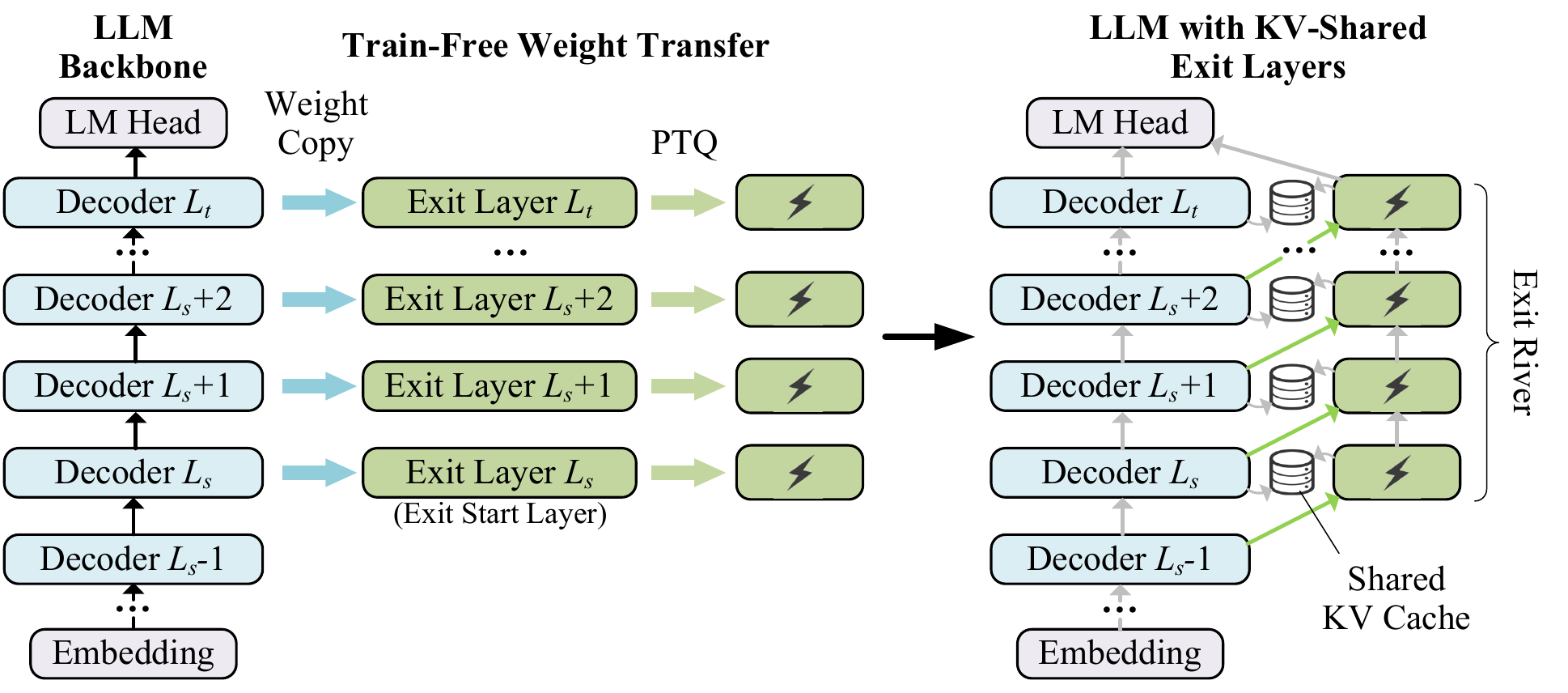}
    \caption{}
	\label{fig:river_llm_a}
	\end{subfigure}
	\begin{subfigure}[c]{0.27\textwidth}
	\centering
	\includegraphics[height=0.16\textheight]{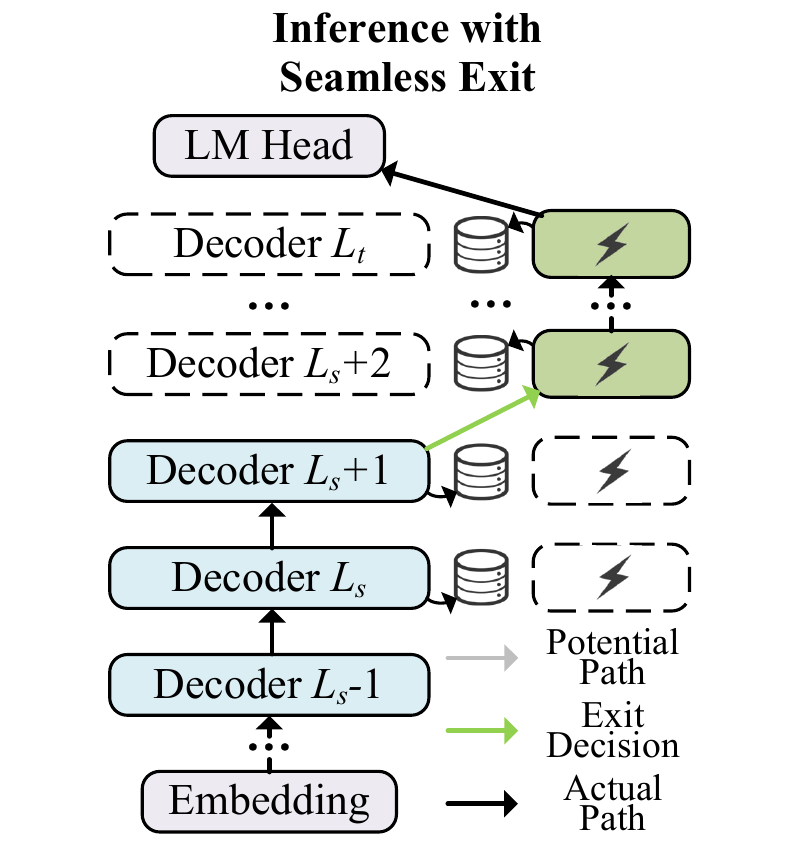}
    \caption{}
	\label{fig:river_llm_b}
	\end{subfigure}
\caption{Seamless exit architecture and inference paradigm: River-LLM. (a) KV-shared exit layer. (b) Inference with seamless exit.}
\label{fig:river_llm}
\end{figure*}

\begin{figure}[!t]
\centering
\centering
	\begin{subfigure}[c]{0.485\textwidth}
    \centering
	\includegraphics[width=0.99\textwidth]{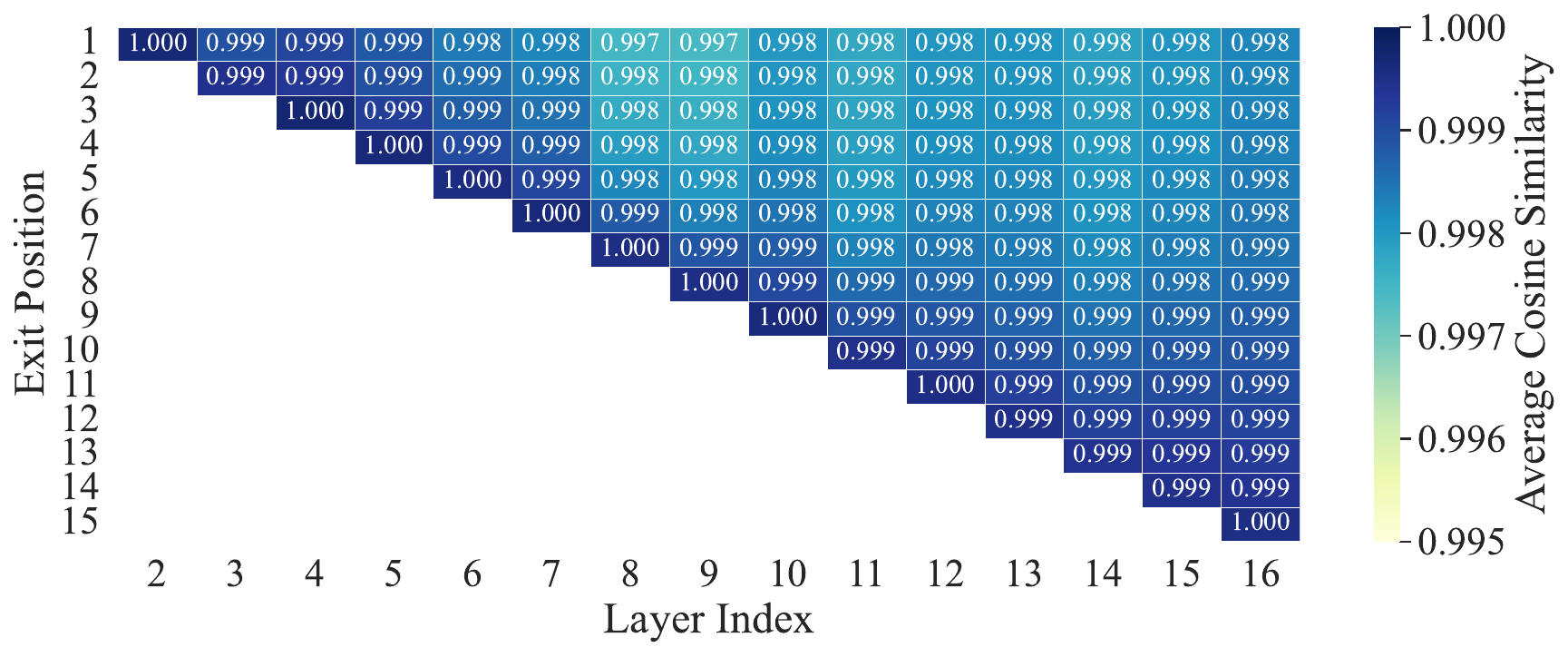}
    \caption{}
	\label{fig:k_heatmap}
	\end{subfigure}
	\begin{subfigure}[c]{0.485\textwidth}
	\centering
	\includegraphics[width=0.99\textwidth]{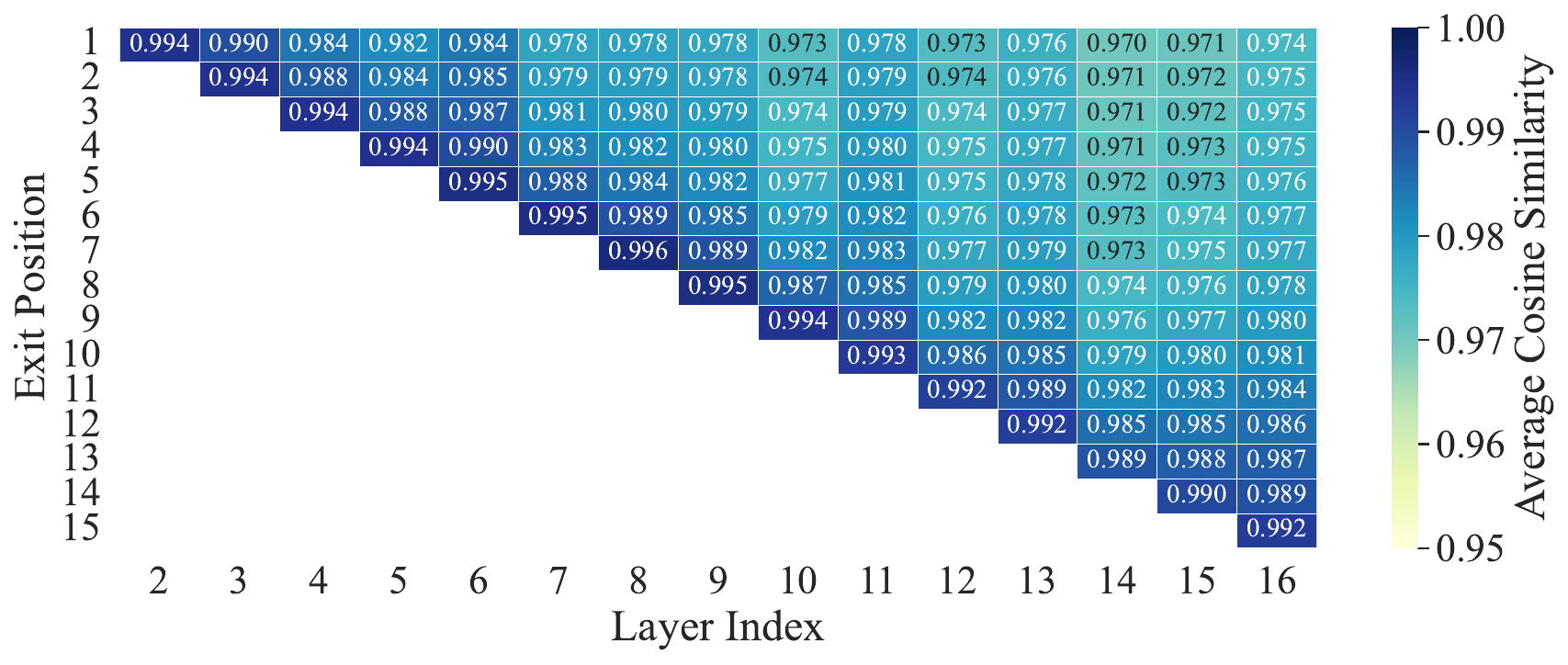}
    \caption{}
	\label{fig:v_heatmap}
	\end{subfigure}
\caption{KV Cache similarity between exit layer and backbone decoder of Llama3.2 1B. (a) Key. (b) Value.}
\label{fig:kV_cos_sim}
\end{figure}

As illustrated in Fig.~\ref{fig:river_llm_a}, the exit layer of River-LLM is designed as a lightweight plug-in specifically for decoder-only Transformers. It adopts a river-like topology, where each exit layer serves as a direct mapping of its corresponding backbone decoder. To ensure Intrinsic KV Integrity, each exit layer is assigned the identical KV Cache addressing scheme as its backbone counterpart. Initially, the exit layers inherit the backbone's architecture and parameters. Subsequently, we apply Post-Training Quantization (PTQ) to the weights of the Attention and Feed-Forward Network blocks within these layers. Specifically, we utilize a 4-bit weight-only quantization (W4A16) scheme, while maintaining the KV Cache in FP16 format to preserve representation density. By leveraging quantization and specialized inference kernels optimized via partial graph compilation, the exit layer allows the hidden states to traverse the "Exit River" with a $2.4\times$ throughput enhancement over the full-precision backbone blocks, while the synthesized KV Cache remains highly consistent with the backbone's native output. According to Fig.~\ref{fig:kV_cos_sim}, regardless of the exit point, the average cosine similarity between the exit KV Cache and the backbone's native KV Cache remains above 0.97. Notably, the KV-Shared mechanism combined with PTQ maintains exceptional Intrinsic KV Integrity without requiring any training; the entire weight transfer process typically concludes within one minute.

\subsection{Inference with Seamless Exit}
\label{sec:inference}

\begin{figure}[t]
\centering
	\begin{subfigure}[b]{0.235\textwidth}
	\includegraphics[height=0.12\textheight]{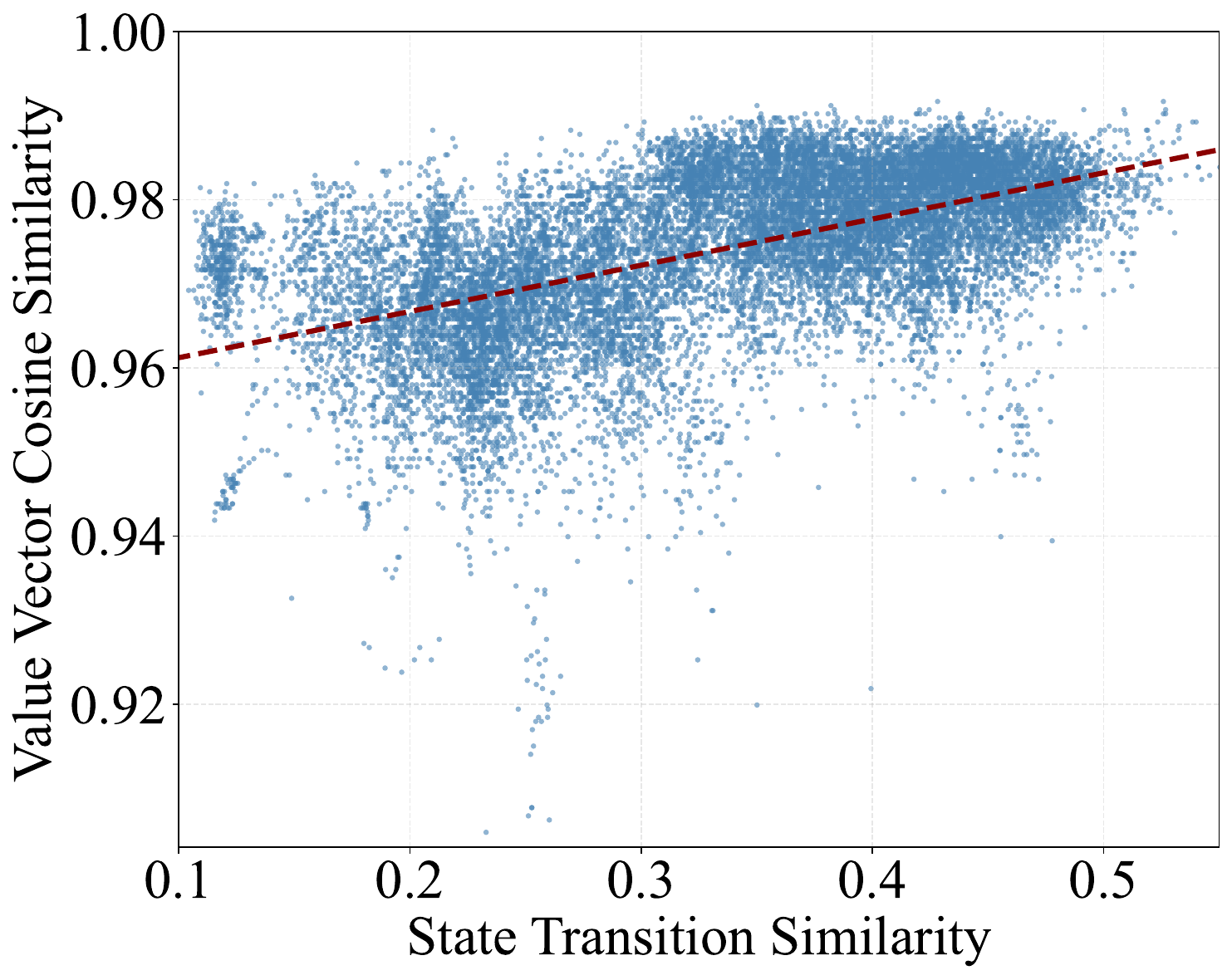}
    \caption{}
	\label{fig:scatter_cos}
	\end{subfigure}
	\hfill
	\begin{subfigure}[b]{0.235\textwidth}
	\centering
	\includegraphics[height=0.12\textheight]{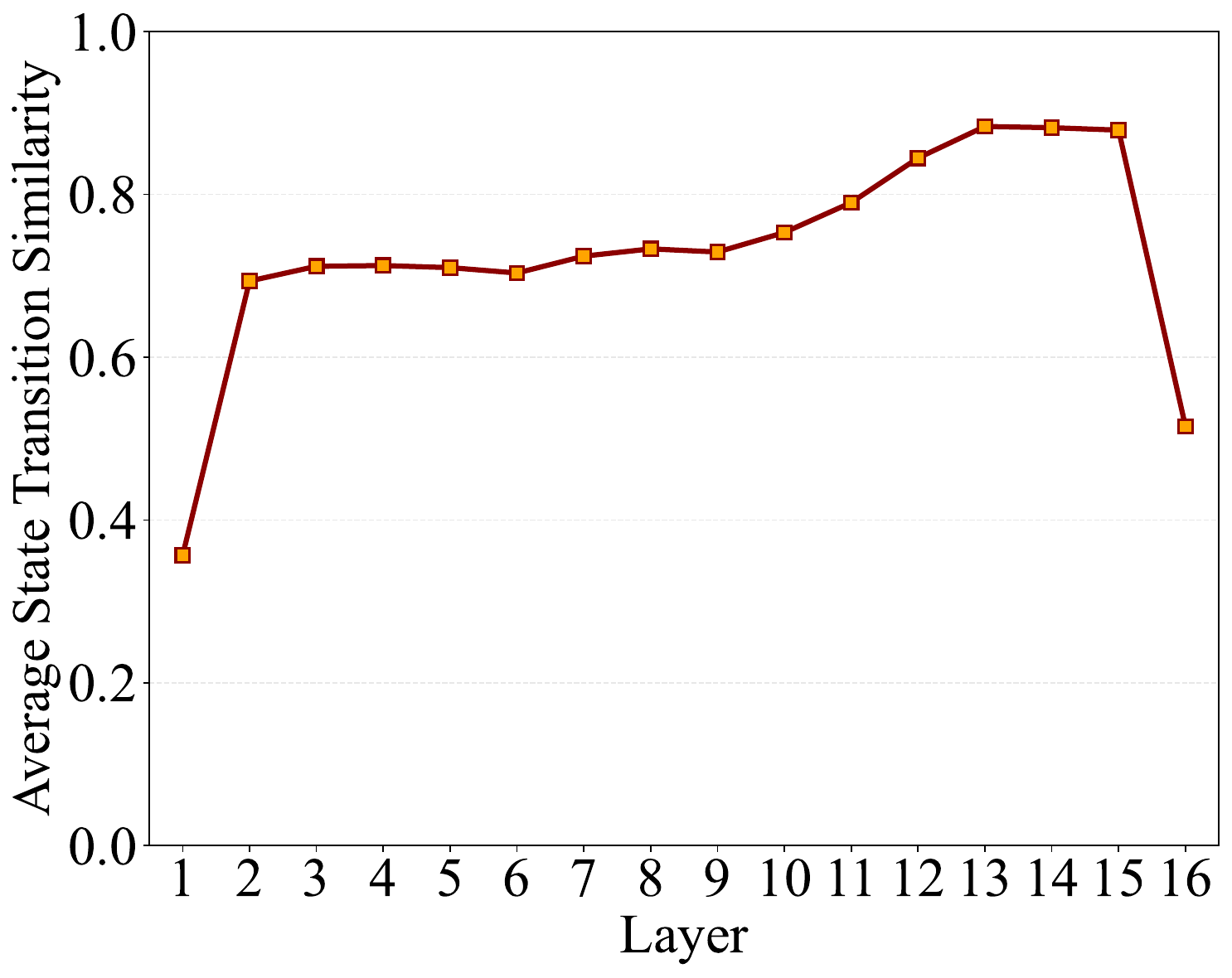}
    \caption{}
	\label{fig:line_cos}
	\end{subfigure}
\label{fig:cos_sim}
\caption{(a) Relationship between first layer state transition similarity and last layer backbone-exit value vector similarity ($r = 0.5536, p < 0.001$). (b) Trend of state transition similarity with the backbone layer index.}
\end{figure}

\begin{figure}[t]
\centering
	\begin{subfigure}[b]{0.235\textwidth}
	\includegraphics[height=0.12\textheight]{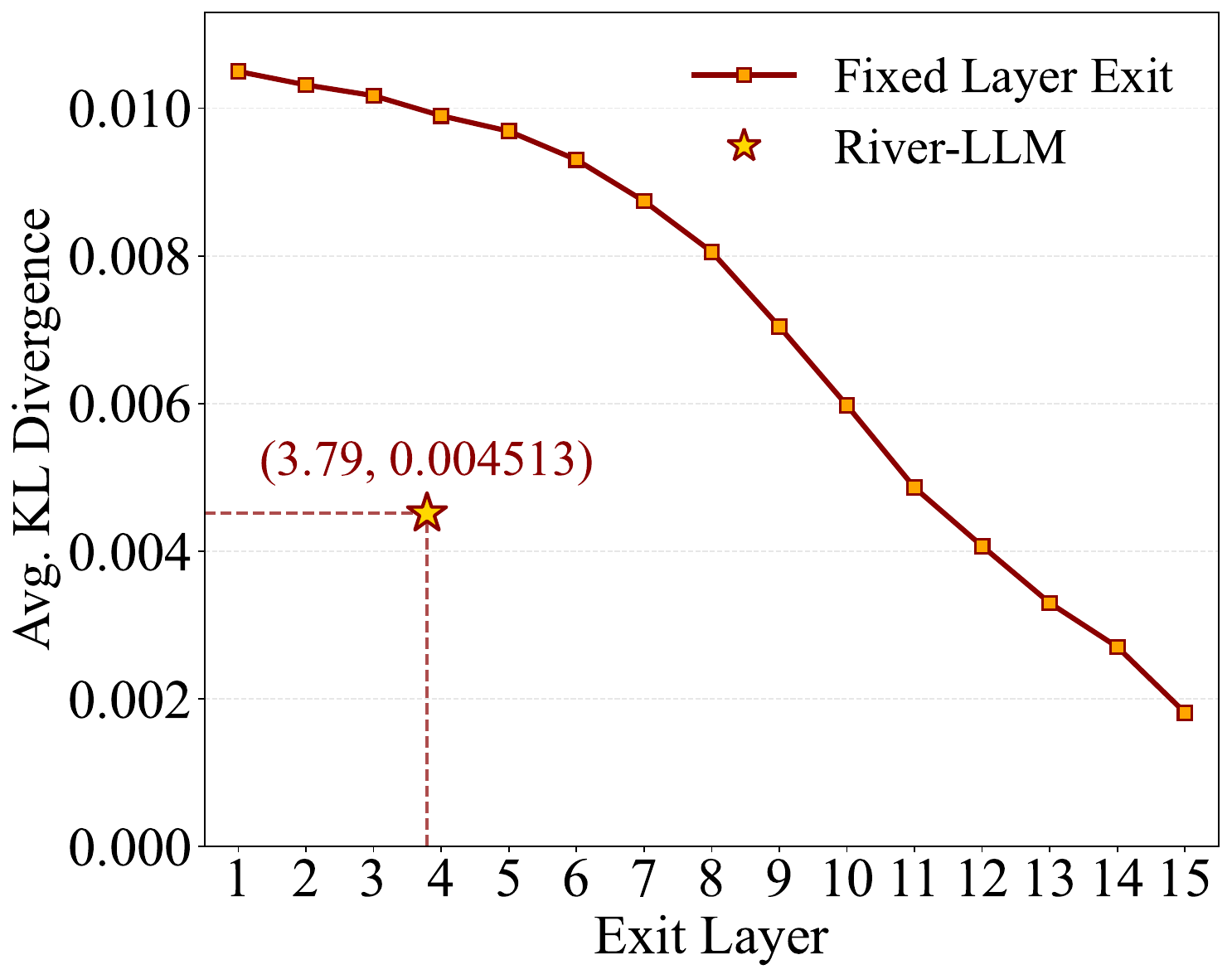}
    \caption{}
	\label{fig:kl}
	\end{subfigure}
	\hfill
	\begin{subfigure}[b]{0.235\textwidth}
	\centering
	\includegraphics[height=0.12\textheight]{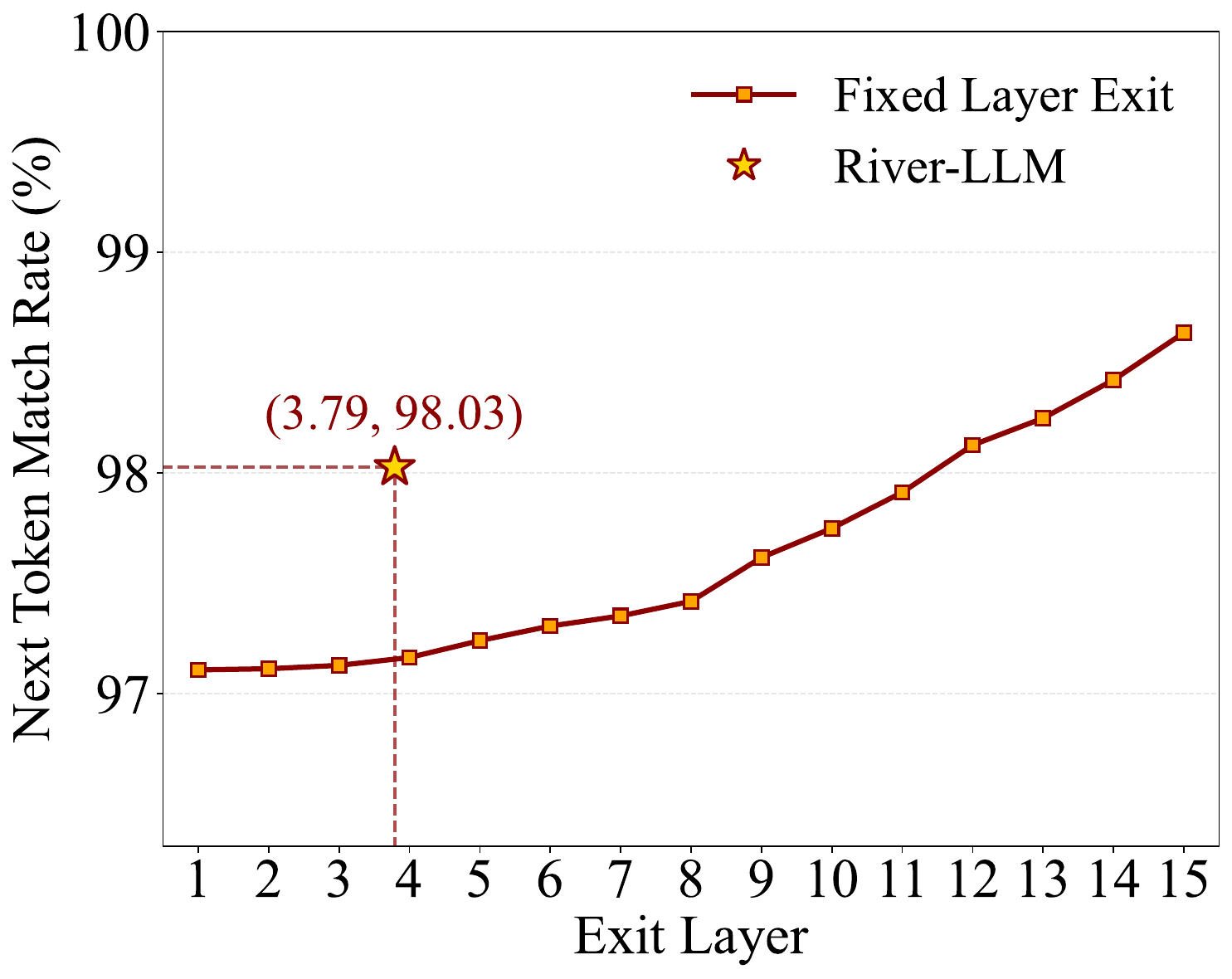}
    \caption{}
	\label{fig:match}
	\end{subfigure}
\caption{(a) Average KL Divergence and (b) next token match rate between backbone output and exit prediction, $\tau=0.5$.}
\label{fig:kl_match}
\end{figure}

To maximize Granular Freedom, River-LLM's exit layers are interconnected in a serial topology starting from a predefined entry layer $L_s$. Each exit layer is logically coupled with its corresponding backbone decoder block. During the autoregressive generation at each decoding step $t$, the $l^{th}$ layer's hidden state $\mathbf{h}_t^{(l)}$ undergoes an exit evaluation prior to entering the next decoder block. As illustrated in Fig.~\ref{fig:river_llm_b}, if the exit criteria are met, the remaining computation is offloaded to the accelerated exit layers, eventually reaching the original LM Head to generate token logits. Exit decision is paramount for maintaining KV Integrity. As evidenced by our analysis in Fig.~\ref{fig:v_heatmap}, we observe a cumulative quantization discrepancy: hidden states passing through multiple 4-bit exit layers exhibit a slight increase in error relative to the backbone. For instance, when exiting at layer 1, the Value vector similarity begins to fluctuate between 0.97 and 0.98 after the 7th exit layer. Notably, we found that the state transition similarity (i.e., the input-output cosine similarity) of early backbone layers serves as a reliable proxy for predicting this cumulative discrepancy. As shown in Fig.~\ref{fig:scatter_cos}, a moderate positive correlation ($r = 0.5536$) exists between the first layer's state transition similarity and the final layer's backbone-exit value similarity. Leveraging this observation, we define the exit decision $\mathcal{D}^{(l)}$ at layer $l$ as:

\begin{small}
\begin{equation}
\mathcal{D}^{(l)} = \mathbb{I} \left( \min_{b \in \mathcal{B}} s_{t,b}^{(l)} > \tau \right),s_{t,b}^{(l)} = \frac{\mathbf{h}_{t,b}^{(l-1) \top} \mathbf{h}_{t,b}^{(l)}}{\|\mathbf{h}_{t,b}^{(l-1)}\| \|\mathbf{h}_{t,b}^{(l)}\|}
\end{equation}
\end{small}
where $s_{t,b}^{(l)}$ is the state transition similarity of generation step $t$ at layer $l$, $\tau$ is the exit threshold and $\mathcal{B}$ is the current batch. Based on Fig.~\ref{fig:line_cos}, cosine similarity generally follows an upward trend across layers, indicating that early and terminal layers exert a more pronounced influence on the hidden states. Except for the last layer, the state transition similarity shows a roughly monotonically increasing trend, which conforms to the objective law of early exit, that is, most of the layers after the exit layer also meet the conditions for early exit.

Building upon this observation, River-LLM facilitates a resource-efficient deployment strategy termed Backbone Offloading. Since the vast majority of tokens terminate their backbone traversal at early stages, the framework can autonomously designate the subsequent, sparsely activated backbone blocks for eviction from primary VRAM. This strategy allows the model to operate within a memory footprint comparable to a fully quantized baseline while ensuring that the Exit River remains resident to provide continuous semantic completion for the few tokens that may require deeper processing.

To harmonize with existing inference frameworks, River-LLM adaptively switches its granularity based on the inference phase:
\begin{itemize}
    \item Prefill Phase: The prompt is processed via Sequence-level Exit, where all tokens exit at a unified depth to maintain the efficiency of parallelized attention kernels.
    \item Generation Phase: Inference switches to Token-level Exit, allowing individual tokens to terminate at their optimal depths to maximize speedup.
\end{itemize}
A significant advantage of River-LLM over full-model quantization is its selective computational fidelity. By allowing "difficult" or high-entropy tokens to traverse the backbone in full precision while offloading "easy" tokens to the Exit River, River-LLM preserves the model's representational robustness in complex scenarios. As illustrated in Fig.~\ref{fig:kl_match}, the quantization discrepancy between the Exit River and the full backbone accumulates as the exit position moves earlier. This accumulation leads to an upward trend in KL divergence between their logit distributions and a corresponding decrease in the next-token match rate. River-LLM leverages this relationship by utilizing the estimated quantization discrepancy to guide precise exit decisions. On the GSM8K benchmark, River-LLM maintains a next-token match rate of 98.03\% and a negligible KL divergence of 0.0045 while achieving an average exit depth of 3.79 layers. This error-aware mechanism ensures that the framework preserves both semantic and KV integrity even when substantial backbone layers are bypassed. Furthermore, by tuning $\tau$, River-LLM facilitates a flexible accuracy-speed trade-off, which we compare against fully quantized baselines in Section~\ref{sec:speed up}.







%% file: latex/evaluation.tex
\section{Evaluation}
\label{sec:eval}
\subsection{Experimental Setup}
\label{sec:setup}
\textbf{Models and Benchmarks}. We evaluate the proposed River-LLM using following representative backbones: Llama3.2 1B, Llama3.1 8B, Phi4-mini and Ministral3 8B. Our evaluation covers a diverse suite of eight benchmarks: 
\begin{itemize}
    \item Common Sense Reasoning: We include BoolQ \cite{clark2019boolq}, HellaSwag \cite{zellers2019hellaswag}, ARC-Challenge \cite{clark2018think}, ARC-Easy, and MMLU \cite{hendrycks2020measuring}. For these tasks, we report accuracy based on loglikelihood ranking.
    \item Long Sequence Generation: We evaluate mathematical reasoning on GSM8K \cite{cobbe2021training} and MATH \cite{hendrycks2021measuring}, alongside code generation on HumanEval \cite{chen2021evaluating}. These benchmarks serve as the primary tasks for measuring practical wall-clock speedup.
\end{itemize}
\textbf{Evaluation Protocol}. All experiments are conducted on an NVIDIA A40 GPU. To ensure robust assessment, we adopt a 5-shot setting for GSM8K and a 4-shot setting for MATH, while all other benchmarks are evaluated in a 0-shot configuration. 


\subsection{Benchmark Accuracy}
\label{sec:acc}

\begin{table*}[t]
\centering
\setlength{\tabcolsep}{5pt}
\begin{footnotesize}
  \centering
  \caption{Evaluation results of River-LLM across diverse benchmarks. The exit position indicates the average number of executed backbone layers. $\tau$ denotes the threshold for the exit decision.}
    \begin{tabular}{lccc|cc|ccc|cc}
    \toprule
            & \multicolumn{5}{c|}{\textbf{Llama3.2 1B (16 layers)}}                & \multicolumn{5}{c}{\textbf{Llama3.1 8B (32 layers)}} \\
    \midrule
    \multicolumn{1}{c}{\multirow{2}[2]{*}{Benchmark}} & \multicolumn{3}{c|}{Accuracy} & \multicolumn{2}{c|}{Exit Position} & \multicolumn{3}{c|}{Accuracy} & \multicolumn{2}{c}{Exit Position} \\
            & \multicolumn{1}{l}{Backbone} & \multicolumn{1}{l}{$\tau=0.5$} & \multicolumn{1}{l|}{$\tau=0.7$} & \multicolumn{1}{l}{$\tau=0.5$} & \multicolumn{1}{l|}{$\tau=0.7$} & \multicolumn{1}{l}{Backbone} & \multicolumn{1}{l}{$\tau=0.5$} & \multicolumn{1}{l|}{$\tau=0.7$} & \multicolumn{1}{l}{$\tau=0.5$} & \multicolumn{1}{l}{$\tau=0.7$} \\
    \midrule
    BoolQ   & 69.4    & \textbf{67.5 } & 69.2    & \textbf{3.90 } & 15.02   & 84.1    & \textbf{83.4 } & 83.4    & \textbf{3.07 } & 9.74  \\
    HellaSwag & 45.1    & \textbf{44.3 } & 44.9    & \textbf{3.71 } & 10.94   & 59.1    & \textbf{58.5 } & 58.6    & \textbf{3.09 } & 8.80  \\
    ARC-c   & 35.7    & \textbf{35.2 } & 35.9    & \textbf{3.24 } & 10.68   & 51.5    & \textbf{50.3 } & 51.2    & \textbf{3.01 } & 8.44  \\
    ARC-e   & 68.4    & \textbf{67.8 } & 67.1    & \textbf{3.24 } & 10.68   & 81.7    & \textbf{82.0 } & 81.4    & \textbf{3.01 } & 8.44  \\
    MMLU    & 46.1    & \textbf{44.3 } & 46.0    & \textbf{4.05 } & 12.70   & 68.0    & \textbf{66.1 } & 67.4    & \textbf{3.04 } & 14.01  \\
    \midrule
    GSM8K   & 33.5    & \textbf{29.3 } & 33.5    & \textbf{3.79 } & 15.05   & 78.2    & \textbf{74.4 } & 75.6    & \textbf{2.96 } & 26.98  \\
    MATH    & 17.8    & \textbf{14.6 } & 17.0    & \textbf{3.56 } & 14.67   & 27.0    & \textbf{26.6 } & 25.7    & \textbf{2.81 } & 13.10  \\
    HumanEval & 25.8    & \textbf{23.2 } & 25.7    & \textbf{2.30 } & 10.05   & 57.3    & \textbf{55.5 } & 57.3    & \textbf{2.16 } & 5.84  \\
    \bottomrule
    \end{tabular}%
  \label{tab:benchmark}%
\end{footnotesize}
\end{table*}%

We first conduct a comprehensive evaluation of River-LLM's accuracy on Llama3.2 1B and Llama3.1 8B across various benchmarks. As demonstrated in Table \ref{tab:benchmark}, under the default threshold ($\tau=0.5$), River-LLM achieves performance competitive with the baseline while executing only 3 to 4 backbone layers on average. This remarkable efficiency is attributed to the KV-Shared Exit River, which maintains KV integrity and enables tokens to exit at extremely early stages without significant semantic degradation. By increasing the threshold to $\tau=0.7$, River-LLM achieves nearly lossless inference. For the 32-layer Llama3.1 8B, the majority of commonsense reasoning and even complex generation tasks can terminate well before the median layer, significantly reducing computational overhead. Interestingly, we observe that River-LLM can even achieve the same level of accuracy as the full-model baseline in specific scenarios, such as HumanEval on Llama3.1 8B. This suggests that by bypassing redundant deeper layers, the Exit River may effectively mitigate cumulative noise or overthinking, thereby enhancing the model's calibration on certain tasks. Consequently, River-LLM proves to be a robust framework that preserves generation quality for complex problems while fully unlocking acceleration potential on simpler tokens.


\subsection{Generation Speedup}
\label{sec:speed up}


\begin{table*}[t]
\begin{footnotesize}
  \centering
  \setlength{\tabcolsep}{3pt}
  \caption{Practical generation speedup and accuracy of River-LLM compared with the backbone and full quantization baselines. Throughput is measured in tokens per second (Tokens/s) on an NVIDIA A40 GPU.}
    \begin{tabular}{cccc|ccc|ccc}
    \toprule
            & \multicolumn{9}{c}{\textbf{Llama3.2 1B}} \\
    \midrule
            & \multicolumn{3}{c|}{\textbf{GSM8K}} & \multicolumn{3}{c|}{\textbf{MATH}} & \multicolumn{3}{c}{\textbf{HumanEval}} \\
            & Backbone & Full Quant. & \textbf{River-LLM} & Backbone & Full Quant. & \textbf{River-LLM} & Backbone & Full Quant. & \textbf{River-LLM} \\
    \midrule
    \textbf{Accuracy} & 33.2    & 25.1    & 29.3    & 17.8    & 12.2    & 14.6    & 25.8    & 20.4    & 23.2  \\
    \textbf{Tokens/s} & 84.5    & 195.5   & 182.9   & 100.5   & 208.8   & 189.2   & 100.4   & 190.6   & 171.8  \\
    \textbf{Speedup} & 1.00$\times$    & 2.31$\times$    & \textbf{2.16$\times$} & 1.00$\times$    & 2.08$\times$    & \textbf{1.88$\times$} & 1.00$\times$    & 1.90$\times$    & \textbf{1.71$\times$} \\
    \midrule
            & \multicolumn{9}{c}{\textbf{Llama3.1 8B}} \\
    \midrule
            & \multicolumn{3}{c|}{\textbf{GSM8K}} & \multicolumn{3}{c|}{\textbf{MATH}} & \multicolumn{3}{c}{\textbf{HumanEval}} \\
            & Backbone & Full Quant. & \textbf{River-LLM} & Backbone & Full Quant. & \textbf{River-LLM} & Backbone & Full Quant. & \textbf{River-LLM} \\
    \midrule
    \textbf{Accuracy} & 78.2   & 69.8    & 74.4    &  27.0  & 23.1    & 26.6    & 57.3    & 50.2    & 55.5  \\
    \textbf{Tokens/s} & 25.3    & 47.5    & 45.0    & 25.1    & 47.6    & 43.1    & 25.2    & 47.7    & 44.7  \\
    \textbf{Speedup} & 1.00$\times$    & 1.88$\times$    & \textbf{1.78$\times$} & 1.00$\times$    & 1.89$\times$    & \textbf{1.72$\times$} & 1.00$\times$    & 1.89$\times$    & \textbf{1.77$\times$} \\
    \midrule
            & \multicolumn{9}{c}{\textbf{Phi4-mini}} \\
    \midrule
            & \multicolumn{3}{c|}{\textbf{GSM8K}} & \multicolumn{3}{c|}{\textbf{MATH}} & \multicolumn{3}{c}{\textbf{HumanEval}} \\
            & Backbone & Full Quant. & \textbf{River-LLM} & Backbone & Full Quant. & \textbf{River-LLM} & Backbone & Full Quant. & \textbf{River-LLM} \\
    \midrule
    \textbf{Accuracy} & 82.1    & 75.9    & 81.0    & 37.4    & 28.8    & 35.1    & 63.4    & 57.9    & 63.1  \\
    \textbf{Tokens/s} & 71.4    & 136.8   & 115.0   & 62.41   & 110.7   & 95.2    & 69.5    & 132.1   & 118.6  \\
    \textbf{Speedup} & 1.00$\times$    & 1.92$\times$    & \textbf{1.61$\times$} & 1.00$\times$    & 1.77$\times$    & \textbf{1.53$\times$} & 1.00$\times$    & 1.90$\times$    & \textbf{1.71$\times$} \\
    \midrule
            & \multicolumn{9}{c}{\textbf{Ministral3 8B}} \\
    \midrule
            & \multicolumn{3}{c|}{\textbf{GSM8K}} & \multicolumn{3}{c|}{\textbf{MATH}} & \multicolumn{3}{c}{\textbf{HumanEval}} \\
            & Backbone & Full Quant. & \textbf{River-LLM} & Backbone & Full Quant. & \textbf{River-LLM} & Backbone & Full Quant. & \textbf{River-LLM} \\
    \midrule
    \textbf{Accuracy} & 84.5    & 84.1    & 84.3    & 48.1    & 46.0    & 46.6    & 29.3    & 20.7    & 26.2  \\
    \textbf{Tokens/s} & 34.9    & 65.0    & 61.8    & 32.0    & 62.7    & 59.1    & 33.7    & 71.3    & 66.6  \\
    \textbf{Speedup} & 1.00$\times$    & 1.86$\times$    & \textbf{1.77$\times$} & 1.00$\times$    & 1.96$\times$    & \textbf{1.85$\times$} & 1.00$\times$    & 2.12$\times$    & \textbf{1.98$\times$} \\
    \bottomrule
    \end{tabular}%
  \label{tab:generation}%
\end{footnotesize}
\end{table*}%

\begin{figure}[t]
\centering
\includegraphics[width=0.43\textwidth]{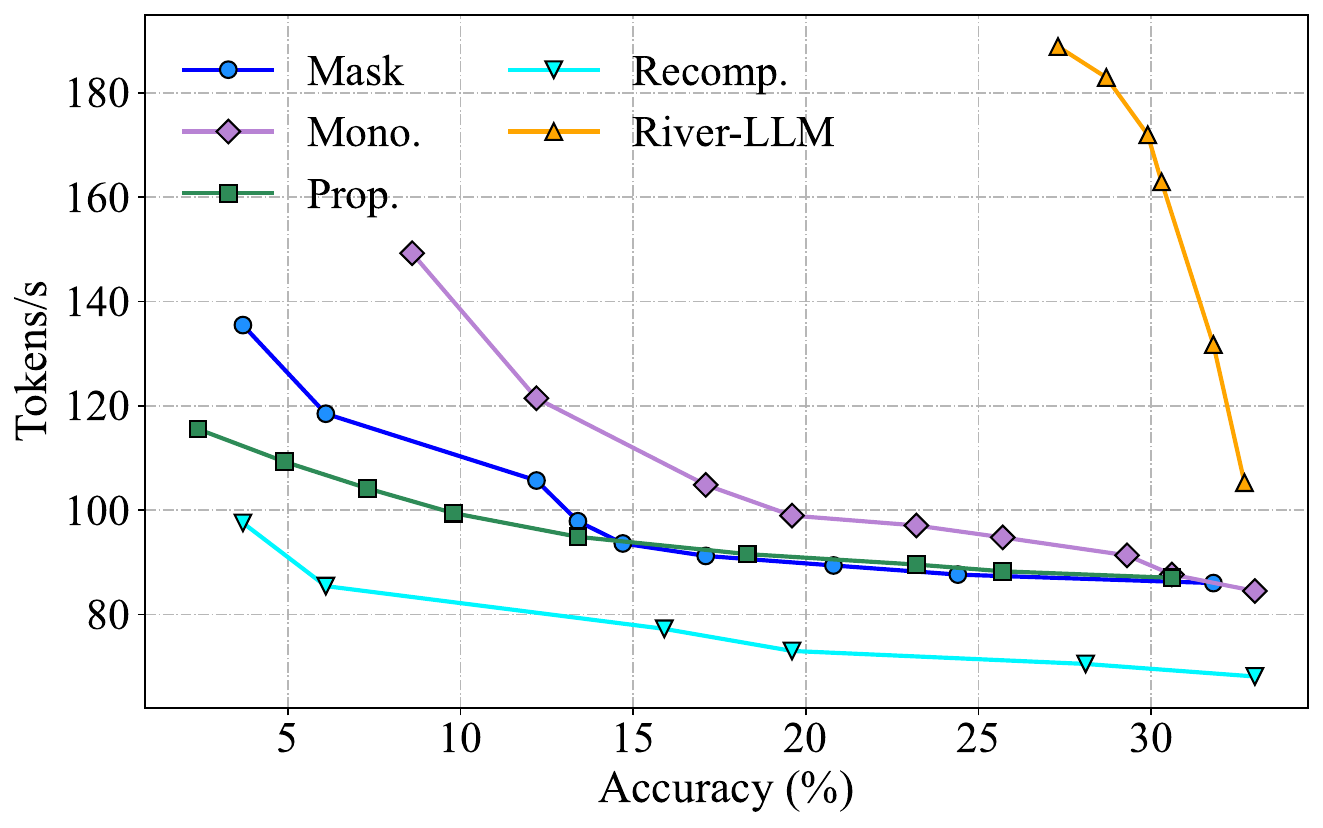}
\caption{Trade-off between generation throughput and GSM8K accuracy on Llama3.2 1B. River-LLM has exceptional accuracy retention capability.}
\label{fig:trade-off}
\end{figure}

Table \ref{tab:generation} reports the practical wall-clock speedup of River-LLM during autoregressive generation on long-sequence benchmarks, including GSM8K, MATH, and HumanEval, with a batch size of 1. We report the average value of 10 tests. For Llama3.2 1B and Llama3.1 8B, we adopt a default exit threshold of 0.5; for Phi4-mini and Ministral3 8B, the threshold is set to 0.9. As a robust baseline for ablation, we evaluate the backbone under Full Quantization (Full Quant.) using the same HQQ framework \cite{hqq} combined with compilation optimization. While static quantization achieves marginally higher peak throughput by applying uniform optimization across all tokens, it incurs significant accuracy degradation due to accumulated precision loss, especially on "difficult" tokens, without the benefit of QAT fine-tuning. In contrast, River-LLM delivers comparable speedups ($\approx 10\%$ lower than Full Quant.) while maintaining near-lossless fidelity to the original backbone's performance.

The versatility of River-LLM is further demonstrated by its ability to navigate the speedup-accuracy trade-off via the exit threshold $\tau$. As illustrated in Fig. \ref{fig:trade-off}, River-LLM consistently outperforms methods with alternative KV strategies, such as masking, recomputation, and state propagation, by a substantial margin. Notably, the performance curve of River-LLM defines a superior Pareto frontier. Unlike traditional methods that suffer from a "cliff-like" drop in accuracy when seeking higher throughput, River-LLM effectively harmonizes the concepts of dynamic depth and KV integrity. This synergy allows the model to selectively skip computation for simpler tokens while preserving the full capacity of the backbone for complex reasoning, thereby achieving an optimal balance between inference efficiency and generation quality.

\subsection{Memory and Latency Overhead}
\label{sec:eval_mem}

\begin{figure}[!t]
\centering
\includegraphics[width=0.47\textwidth]{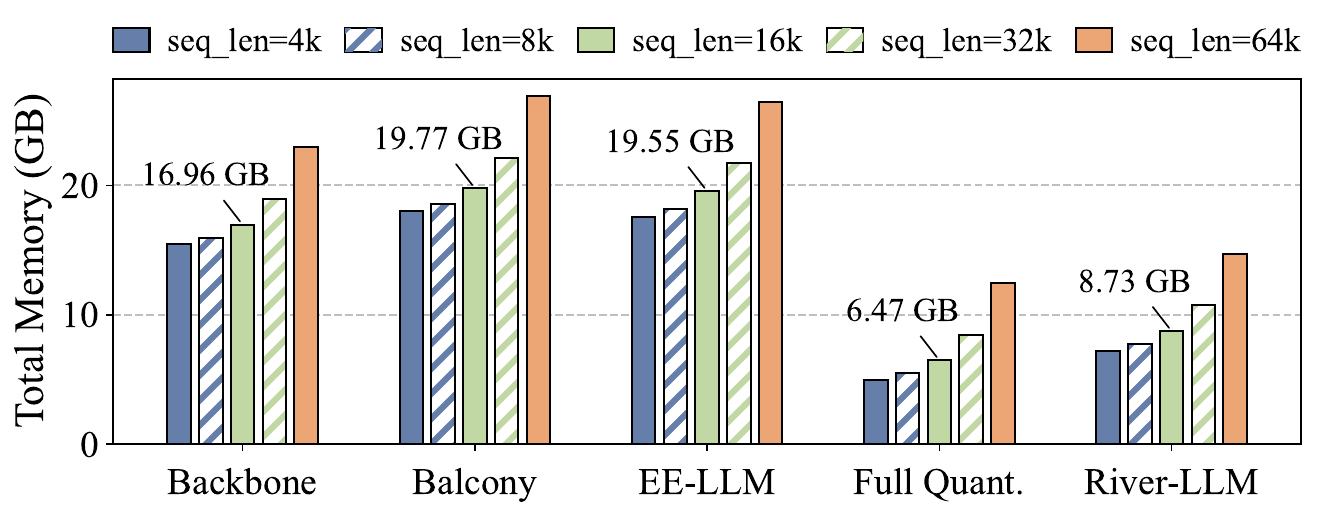}
\caption{Peak GPU memory usage of Llama3.1 8B with different methods, $\mathtt{batch\_size} = 1$.}
\label{fig:mem}
\end{figure}
To evaluate the deployment efficiency of River-LLM, we analyze its GPU memory consumption across varying sequence lengths, ranging from 4K to 64K tokens. As illustrated in Fig.~\ref{fig:mem}, prior early exit frameworks such as Balcony and EE-LLM incur substantial memory overhead that regularly exceeds the original backbone, primarily due to the retention of additional exit layer parameters or duplicated KV cache states. In contrast, by leveraging the single shared KV mechanism alongside the backbone offloading strategy, River-LLM dramatically reduces memory requirements across all context lengths. Its total memory footprint remains consistently and significantly lower than both the backbone and existing early exit baselines, approaching the memory efficiency of a fully quantized model while still preserving selective high-precision inference capabilities.

Beyond memory efficiency, River-LLM introduces negligible latency overhead during inference. The exit criterion relies on computing the state transition similarity between consecutive hidden states, which, for a hidden dimension $d$, incurs a time complexity of $\mathcal{O}(d)$. Empirical profiling corroborates this efficiency: the exit decision logic executes in approximately 100 microseconds on Llama3.1 8B, accounting for a mere 0.0688\% of the total per-token inference time. This confirms that River-LLM’s seamless exit mechanism delivers substantial speedup with minimal overhead.








%% file: latex/conclusion.tex
\section{Conclusion}
\label{sec:conclusion}
In this paper, we address the KV Cache Absence problem, a fundamental bottleneck that hinders the practical efficiency of Early Exit in decoder-only Large Language Models. Our empirical analysis demonstrates that existing remedies, ranging from recomputation to masking, fail to bridge the gap between theoretical layer reduction and actual wall-clock speedup due to their substantial latency overhead or precision degradation.
To overcome these limitations, we introduce River-LLM, a training-free framework designed for seamless token-level Early Exit. By constructing a lightweight KV-Shared Exit River, River-LLM enables the backbone’s missing KV cache to be naturally generated and preserved as an intrinsic byproduct of the exit process. Furthermore, we utilize state transition similarity within decoder blocks to guide precise exit decisions, ensuring high fidelity to the original backbone's output.
Extensive evaluations on diverse benchmarks demonstrate that River-LLM achieves 1.71$\times$ to 2.16$\times$ practical speedup while maintaining near-lossless generation quality. Compared to prior dynamic inference methods, River-LLM defines a superior Pareto frontier, offering a flexible accuracy-speed trade-off without the need for additional training or fine-tuning. In summary, River-LLM provides a robust and scalable solution for efficient LLM inference, proving that maintaining KV integrity is the key to fully unlocking the token-level exit potential.

%% file: latex/appendix.tex
\section{Appendix}
\label{sec:appendix}

\subsection{Impact of Different Quantization Backends in Exit River}
\label{sec:quant_method}

To demonstrate the framework’s architecture-agnostic design, we evaluate the performance of River-LLM when integrated with various quantization backends for the Exit River path. As a dynamic inference framework, River-LLM can incorporate any mature compression technique that preserves Intrinsic KV Integrity during the exit process. Table~\ref{tab:awq} summarizes the results for Llama-3.1-8B using HQQ (the default method used by River-LLM) and AWQ as alternative quantization schemes.

\begin{table}[t]
\setlength{\tabcolsep}{3pt}
\begin{footnotesize}
  \centering
  \caption{Different Exit River Quantization on Llama3.1 8B.}
    \begin{tabular}{cccc}
    \toprule
    \textbf{Quant. Method} & \textbf{Exit Threshold} & \multicolumn{1}{p{4.04em}}{\textbf{Token/s}} & \multicolumn{1}{p{4.04em}}{\textbf{GSM8K Score}} \\
    \midrule
    \textbf{Backbone} & {N/A} & {25.3} & {78.2} \\
    {\textbf{River+HQQ}} & \multicolumn{1}{c}{{0.5}} & {45} & {74.4} \\
    \textbf{HQQ} & {full quant} & {47.5} & {69.8} \\
    {\textbf{River+AWQ}} & \multicolumn{1}{c}{{0.5}} & {46.9} & {77.3} \\
    \textbf{AWQ} & {full quant} & {48.1} & {76.2} \\
    \bottomrule
    \end{tabular}%
  \label{tab:awq}%
\end{footnotesize}
\end{table}%

The results indicate that advancements in static quantization directly translate to improved performance within the River-LLM framework. When the Exit River is upgraded from the default HQQ to the more advanced AWQ, the model achieves higher accuracy across reasoning tasks. Specifically, the River + AWQ configuration achieves a GSM8K score of 77.3, outperforming the static AWQ baseline (76.2) while maintaining competitive throughput.

This precision gain demonstrates that River-LLM acts as a complementary path-level optimization for operator-level quantization. By adaptively routing high-entropy tokens through the full-precision backbone, River-LLM effectively mitigates the cumulative quantization noise inherent in static low-bit models. These findings confirm that River-LLM can enhance the accuracy of existing lightweight methods with a negligible impact on acceleration efficiency.

\subsection{Benchmark results of Phi4-mini and Ministral3 8B}
\label{sec:phi_ministral}

\begin{table*}[t]
\centering
\setlength{\tabcolsep}{5pt}
\begin{footnotesize}
  \centering
  \caption{Additional evaluation results of River-LLM on Phi4-mini and Ministral3 8B across diverse benchmarks.}
    \begin{tabular}{lccc|cc|ccc|cc}
    \toprule
            & \multicolumn{5}{c}{\textbf{Phi4-mini (32 layers)}}          & \multicolumn{5}{c}{\textbf{Ministral3 8B (34 layers)}} \\
    \midrule
    \multicolumn{1}{c}{\multirow{2}[2]{*}{Benchmark}} & \multicolumn{3}{c|}{Accuracy} & \multicolumn{2}{c|}{Exit Position} & \multicolumn{3}{c|}{Accuracy} & \multicolumn{2}{c}{Exit Position} \\
            & Backbone & \multicolumn{1}{l}{$\tau=0.8$} & \multicolumn{1}{l|}{$\tau=0.9$} & \multicolumn{1}{l}{$\tau=0.8$} & \multicolumn{1}{l|}{$\tau=0.9$} & Backbone & \multicolumn{1}{l}{$\tau=0.8$} & \multicolumn{1}{l|}{$\tau=0.9$} & \multicolumn{1}{l}{$\tau=0.8$} & \multicolumn{1}{l}{$\tau=0.9$} \\
    \midrule
    BoolQ   & 84.3    & 84.1    & \textbf{84.3 } & 2.35    & \textbf{4.93 } & 85.9    & 85.2    & \textbf{85.4 } & 2.04    & \textbf{3.83 } \\
    HellaSwag & 54.4    & 53.7    & \textbf{53.9 } & 2.05    & \textbf{4.14 } & 58.5    & 58.2    & \textbf{58.2 } & 2.00    & \textbf{2.38 } \\
    ARC-c   & 57.1    & 54.0    & \textbf{54.4 } & 2.02    & \textbf{3.45 } & 62.3    & 62.2    & \textbf{62.2 } & 2.01    & \textbf{2.11 } \\
    ARC-e   & 82.8    & 81.3    & \textbf{81.3 } & 2.02    & \textbf{3.45 } & 86.5    & 85.9    & \textbf{85.9 } & 2.01    & \textbf{2.11 } \\
    MMLU    & 66.7    & 64.1    & \textbf{65.4 } & 2.07    & \textbf{4.48 } & 73.1    & 72.7    & \textbf{72.7 } & 2.01    & \textbf{2.76 } \\
    \midrule
    GSM8K   & 82.1    & 79.2    & \textbf{81.0 } & 6.10    & \textbf{13.79 } & 84.5    & 84.2    & \textbf{84.3 } & 6.16    & \textbf{6.70 } \\
    MATH    & 37.4    & 33.1    & \textbf{35.1 } & 5.77    & \textbf{13.69 } & 48.1    & 46.4    & \textbf{46.6 } & 5.68    & \textbf{6.18 } \\
    HumanEval & 63.4    & 62.8    & \textbf{63.1 } & 2.55    & \textbf{7.72 } & 29.3    & 22.6    & \textbf{26.2 } & 2.51    & \textbf{7.02 } \\
    \bottomrule
    \end{tabular}%
  \label{tab:benchmark_2}%
\end{footnotesize}
\end{table*}%

Table \ref{tab:benchmark_2} presents the additional benchmark evaluation of River-LLM on the Phi4-mini and Ministral3 8B architectures. Consistent with the findings on Llama models, the framework preserves generation quality across diverse benchmarks while reducing the overall computational depth. A key observation from these experiments is that newer model architectures demonstrate stronger early-layer semantic understanding, causing tokens to converge to stable representations at much shallower layers. Consequently, we applied stricter exit thresholds ($\tau=0.8$ and $\tau=0.9$) for these models to simulate a near-lossless acceleration scenario. Under these elevated thresholds, River-LLM maintains high-fidelity performance across both common sense and complex reasoning tasks while still bypassing a significant portion of the backbone, further confirming the architecture-agnostic scalability of the KV-shared exit mechanism.

\subsection{Detailed Memory Usage of River-LLM and Other Baselines}
\label{sec:mem}

Table \ref{tab:mem} provides a granular decomposition of peak GPU memory consumption, partitioned into model parameters, KV cache, and temporary activations. A key observation is the scalability of the KV-shared architecture in long-sequence scenarios. While representative early exit frameworks like Balcony and EE-LLM require auxiliary KV cache sets for their respective exit layers or recomputation states (resulting in higher GPU memory consumption than the original backbone) River-LLM maintains a single set of KV caches identical to the backbone baseline. Consequently, as the context length increases to 64K, River-LLM avoids the memory inflation inherent in competing dynamic depth strategies.

Regarding parameter footprint, the Backbone-Offloading mode demonstrates significant efficiency gains. By offloading under-utilized backbone layers while retaining the Exit River, the parameter memory is reduced from 14.96 GB to 6.73 GB. This configuration allows the total memory footprint to approach that of a static 4-bit quantization baseline while selectively preserving high-precision execution for the critical initial layers of the model. These results confirm that River-LLM offers a flexible deployment strategy where the trade-off between memory capacity and generation fidelity is explicitly manageable.

\begin{table}[t]
\setlength{\tabcolsep}{3.9pt}
  \centering
  \begin{footnotesize}
  \caption{Peak GPU Memory Usage of Different Methods for Llama3.1 8B, $\mathtt{batch\_size} = 1$.}
    \begin{tabular}{cccccc}
    \toprule
    \multicolumn{1}{c}{\textbf{Method}} & \multicolumn{1}{p{3.25em}}{\textbf{Context Length}} & \multicolumn{1}{p{2.25em}}{\textbf{Para-meters (GB)}} & \multicolumn{1}{p{2.54em}}{\textbf{KV Cache (MB)}} & \multicolumn{1}{p{2.54em}}{\textbf{Temp Act. (MB)}} & \multicolumn{1}{p{2.54em}}{\textbf{Total Mem. (GB)}} \\
    \midrule
    \multirow{5}[2]{*}{\textbf{Backbone}} & 4,096   & 14.96   & 512     & 1.25    & 15.46 \\
            & 8,192   & 14.96   & 1024    & 1.25    & 15.96 \\
            & 16,384  & 14.96   & 2048    & 1.25    & 16.96 \\
            & 32,768  & 14.96   & 4096    & 1.25    & 18.96 \\
            & 65,536  & 14.96   & 8192    & 1.25    & 22.96 \\
    \midrule
    \multicolumn{1}{c}{\multirow{5}[2]{*}{\textbf{Balcony}}} & 4,096   & 17.4    & 608     & 1.25    & 17.99 \\
            & 8,192   & 17.4    & 1216    & 1.25    & 18.58 \\
            & 16,384  & 17.4    & 2432    & 1.25    & 19.77 \\
            & 32,768  & 17.4    & 4864    & 1.25    & 22.15 \\
            & 65,536  & 17.4    & 9728    & 1.25    & 26.90 \\
    \midrule
    \multicolumn{1}{c}{\multirow{5}[2]{*}{\textbf{EE-LLM}}} & 4,096   & 16.98   & 601     & 1.25    & 17.58 \\
            & 8,192   & 16.98   & 1202    & 1.25    & 18.17 \\
            & 16,384  & 16.98   & 2397    & 1.25    & 19.55 \\
            & 32,768  & 16.98   & 4836    & 1.25    & 21.72 \\
            & 65,536  & 16.98   & 9701    & 1.25    & 26.47 \\
    \midrule
    \multirow{5}[2]{*}{\textbf{Full Quantize}} & 4,096   & 4.47    & 512     & 1.25    & 4.97 \\
            & 8,192   & 4.47    & 1024    & 1.25    & 5.47 \\
            & 16,384  & 4.47    & 2048    & 1.25    & 6.47 \\
            & 32,768  & 4.47    & 4096    & 1.25    & 8.47 \\
            & 65,536  & 4.47    & 8192    & 1.25    & 12.47 \\
    \midrule
            & 4,096   & 6.73    & 512     & 1.25    & 7.23 \\
    {} & 8,192   & 6.73    & 1024    & 1.25    & 7.73 \\
    {\textbf{River-LLM}} & 16,384  & 6.73    & 2048    & 1.25    & 8.73 \\
    {} & 32,768  & 6.73    & 4096    & 1.25    & 10.73 \\
            & 65,536  & 6.73    & 8192    & 1.25    & 14.73 \\
    \bottomrule
    \end{tabular}%
  \label{tab:mem}%
\end{footnotesize}
\end{table}%

\subsection{Additional Discussion: River-LLM v.s. Mixed-Precision LLM}
\label{sec:mixed-precision}

Mixed-precision quantization has emerged as an orthogonal paradigm to Early Exit for optimizing LLM inference efficiency. Existing research typically falls into two categories: static and dynamic assignment. Static frameworks, such as SHMQ \cite{SHMQ}, pre-calculate hierarchical bit-width configurations using Hessian matrices to reflect model-inherent sensitivity, InfoQ \cite{infoq} determines static precision assignments by measuring the impact of layer-wise quantization on global mutual information flow and solving an integer linear programming problem prior to deployment. Conversely, dynamic approaches emphasize that layer importance shifts during generation. PMPD \cite{PMPD} introduces a progressive decoding schedule that reduces precision as the sequence length increases. Representing the current state-of-the-art in token-wise adaptation, DP-LLM \cite{DP-LLM} switches bit-widths at each decoding step based on the statistical properties of input activations, aiming to maximize throughput by navigating the sensitivity of different layers across the generation process.

River-LLM differs from mixed precision frameworks like DP-LLM in its optimization objective, deployment complexity, and accuracy retention. Unlike DP-LLM, which necessitates a calibration phase to train linear regressors for precision assignment, River-LLM is entirely training-free and introduces a nearly negligible decision overhead of approximately 0.0688\%. Furthermore, the two paradigms represent divergent acceleration trajectories: while mixed-precision methods focus on aggressive hardware throughput via low-bit optimization, River-LLM prioritizes preserving the original backbone’s reasoning fidelity. Our comparative analysis highlights this distinction in performance stability. On the GSM8K benchmark, River-LLM on Llama3.1 8B maintains a high-fidelity accuracy range of 74.4\%–78.2\% across a target bit range of 4.00–16.00. In contrast, DP-LLM exhibits a more pronounced performance drop on Llama3 8B, with accuracy fluctuating between 36.7\% and 46.9\% within a lower target bit range of 3.25–4.75. These results indicate that River-LLM offers a superior Pareto frontier for applications requiring near-lossless generation quality, whereas mixed-precision methods are optimized for extreme bit-reduction scenarios that may impose specific hardware requirements. River-LLM thus provides a lightweight, depth-aware mechanism that complements operator-level quantization by adaptively bridging the gap between high-fidelity backbone execution and accelerated inference.